\documentclass[11pt,a4paper]{article}
\usepackage[hyperref]{emnlpijcnlp2019}
\usepackage{times}
\usepackage{amsmath}
\usepackage{amsfonts}
\usepackage{amssymb}
\usepackage{latexsym}
\usepackage{graphicx}
\usepackage{caption}
\usepackage{subcaption}
\usepackage{multirow}
\usepackage{url}

\aclfinalcopy 

\title{Feature2Vec: Distributional semantic modelling of human property knowledge}

\author{
	Steven Derby
	\ \ \ \ \ \ \ \ \ \
	Paul Miller
	\ \ \ \ \ \ \ \ \ \
    Barry Devereux
    \\\\
  Queen's University Belfast, Belfast, United Kingdom \\
  {\{\tt sderby02, \tt p.miller,  \tt b.devereux\}@qub.ac.uk}
}

\date{}

\begin{document}

\maketitle

\begin{abstract}
Feature norm datasets of human conceptual knowledge, collected in surveys of human volunteers, yield highly interpretable models of word meaning and play an important role in neurolinguistic research on semantic cognition. However, these datasets are limited in size due to practical obstacles associated with exhaustively listing properties for a large number of words. In contrast, the development of distributional modelling techniques and the availability of vast text corpora have allowed researchers to construct effective vector space models of word meaning over large lexicons. However, this comes at the cost of interpretable, human-like information about word meaning. We propose a method for mapping human property knowledge onto a distributional semantic space, which adapts the \emph{word2vec} architecture to the task of modelling concept features. Our approach gives a measure of concept and feature affinity in a single semantic space, which makes for easy and efficient ranking of candidate human-derived semantic properties for arbitrary words. We compare our model with a previous approach, and show that it performs better on several evaluation tasks. Finally, we discuss how our method could be used to develop efficient sampling techniques to extend existing feature norm datasets in a reliable way. 
\end{abstract}

\section{Introduction}
Distributional semantic modelling of word meaning has become a popular method for building pretrained lexical representations for downstream Natural Language Processing (NLP) tasks \cite{baroni2010distributional, mikolov2013distributed, pennington2014glove, levy2014dependency, bojanowski2017enriching}. In this approach, meaning is encoded in a dense vector space model, such that words (or concepts) that have vector representations that are spatially close together are similar in meaning. A criticism of these real-valued embedding vectors is the opaqueness of their representational dimensions and their lack of cognitive plausibility and interpretability \cite{murphy2012learning,senel2018}. In contrast, human conceptual property knowledge is often modelled in terms of relatively sparse and interpretable vectors, based on verbalizable, human-elicited features collected in property knowledge surveys \cite{mcrae2005semantic, devereux2014centre}. However, gathering and collating human-elicited property knowledge for concepts is very labour intensive, limiting both the number of words  for which  a rich feature set can be gathered, as well as the completeness of the feature listings for each word. Neural embedding models, on the other hand, learn from large corpora of text in an unsupervised fashion, allowing very detailed, high-dimensional semantic models to be constructed for a very large number of words. In this paper, we propose \textit{Feature2Vec}, a computational framework that combines information from human-elicited property knowledge and information from distributional word embeddings, allowing us to exploit the strengths and advantages of both approaches.  Feature2Vec maps human property norms onto a pretrained vector space model of word meaning. The embedding of feature-based information in the pretrained embedding space makes it possible to rank the relevance of features using cosine similarity, and we demonstrate how simple composition of features can be used to approximate concept vectors. 

\section{Related Work} 
Several property-listing studies have been conducted with human participants in order to build \emph{property norms} -- datasets of normalized human-verbalizable feature listings for lexical concepts \cite{mcrae2005semantic, devereux2014centre}. One use of feature norms is to critically examine distributional semantic models on their ability to encode grounded, human-elicited semantic knowledge. For example, \newcite{rubinstein2015well} demonstrated that state-of-the-art distributional semantic models fail to predict attributive properties of concept words (e.g.~the properties \emph{is-red} and \emph{is-round} for the word \emph{apple}) as accurately as taxonomic properties (e.g.~\emph{is-a-fruit}). Similarly, \newcite{sommerauer2018firearms} investigated the types of semantic knowledge encoded within pretrained word embeddings, concluding that some properties cannot be learned by supervised classifiers.   \newcite{collell2016image} compared linguistic and visual representations of object concepts on their ability to represent different types of property knowledge. Research has shown that state-of-the-art distributional semantic models built from text corpora fail to capture important aspects of meaning related to grounded perceptual information, as this kind of information is not adequately represented in the statistical regularities of text data \cite{lucy2017distributional, kelly2014}. Motivated by these issues, \newcite{silberer2017grounding} constructed multimodal semantic models from text and image data, with the goal of grounding word meaning using visual attributes. More recently, \newcite{derby2018using} built similar models with the added constraint of sparsity, demonstrating that sparse multimodal vectors provide a more faithful representation of human semantic representations. Finally, the work that most resembles ours is that of \newcite{fagarasan2015distributional}, who use Partial Least Squares Regression (PLSR) to learn a mapping from a word embedding model onto specific conceptual properties. 
Concurrent work recently undertaken by \newcite{li2019mapping} replaces the PLSR model with a feedforward neural network. In our work, we instead map property knowledge directly into vector space models of word meaning, rather than learning a supervised predictive function from concept embedding dimensions to feature terms.

\section{Method}
We make primary comparison with the work of \newcite{fagarasan2015distributional}, although their approach differs from ours in that they map from an embedding space onto the feature space, while we learn a mapping from the feature domain onto the embedding space. We outline both methods below.

\subsection{Distributional Semantic Models}
For our experiments, we make use of the pretrained GloVe embeddings \cite{pennington2014glove} provided in the \textit{Spacy}\footnote{https://spacy.io/models/en\#en\_core\_web\_lg} package trained on the Common Crawl\footnote{http://commoncrawl.org/}. The GloVe model includes $685,000$ tokens with embedding vectors of dimension $300$, providing excellent lexical coverage with a rich set of semantic representations.

In our analyses we use both the McRae property norms \cite{mcrae2005semantic}, which contain $541$ concepts and $2526$ features, and the CSLB norms \cite{devereux2014centre} which have $638$ concepts with $2725$ features. For both sets of norms, a feature is listed for a concept if it has been elicited by five or more participants in the property norming study. The number of participants listing a given feature for a given concept name is termed the \emph{production frequency} for that concept$\times$feature pair. This gives sparse production frequency vectors for each concept over all the features in the norms.

\begin{figure*}[h]
	\centering
	\includegraphics[width = \textwidth]{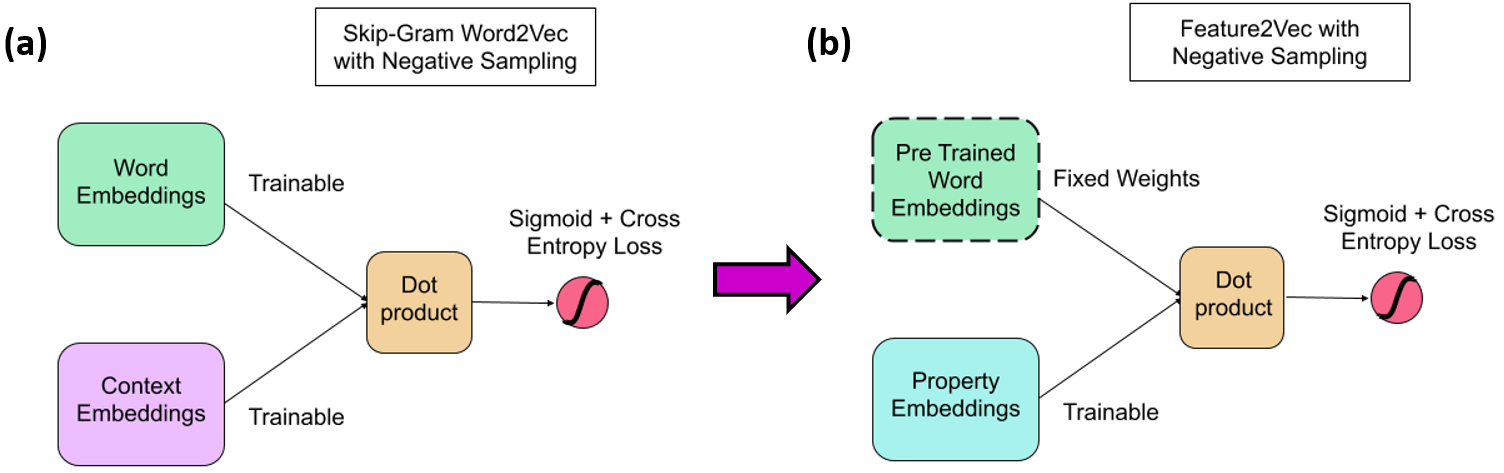}
	\caption{Diagram illustrating the relationship of \emph{Feature2Vec} to the standard Skip-Gram Word2Vec architecture. \textbf{(a)} The standard Skip-Gram Word2Vec architecture with negative sampling. Word and context embeddings are randomly initialized, and both sets of embeddings are trainable. \textbf{(b)} Our Feature2Vec architecture, which uses pretrained word embeddings that are fixed during training. The network learns semantic feature embeddings for feature labels in property norm datasets instead of context embeddings.}
	\label{w2v}
\end{figure*}

\subsection{Partial Least Square Regression (PLSR)}
\newcite{fagarasan2015distributional} used partial least squares regression (PLSR) to map between the GloVe embedding space and property norm vectors. Suppose we have two real-valued matrices $G \in \mathbb{R}^{n \times m}$ and $F \in \mathbb{R}^{n \times k}$. In this context, $G$ and $F$ represent GloVe embedding vectors and property norm feature vectors, respectively. For $n$ available concept words, $G$ is a matrix which consists of stacked pretrained embeddings from GloVe and $F$ is the (sparse) matrix of production frequencies for each concept$\times$feature pair. $G$ and $F$ share the same row indexing for concept words. For a new dimension size $p \in \mathbb{N}$, a partial least squared regression learns two new subspaces with dimensions $n \times p$, which have maximal covariance between them.  The algorithm solves this problem by learning a mapping from the matrix $G$ onto $F$, similar to a regression model.  The fitted regression model thus provides a framework for predicting vectors in the feature space from vectors in the embedding space.

In this work, we use the PLSR approach as a baseline for our model. In implementing PLSR, we set the intermediate dimension size to $50$, following \newcite{fagarasan2015distributional}. 
We also build a PLSR model using $120$ dimensions, which in preliminary experimentation we found gave the best performance from a range of values tested.

\subsection{Skip-Gram Word2Vec}
\newcite{mikolov2013distributed} proposed learning word embeddings using a predictive neural-network approach. In particular, the skip-gram implementation with negative sampling mines word co-occurrences within a text window, and the network must learn to predict the surrounding context from the target word. More specifically, for a vocabulary $V$, two sets of embeddings are learned through gradient decent, one for target embeddings and one for context embeddings. Given a target word $w \in V$ and a context word $c \in V$ in its window, the network calculates the dot product for the embeddings for $w$ and $v$ and a sigmoid activation is applied to the output (Fig.~1(a)). Negative samples are also generated for training, where the context is not in the target word's window. Let $(w,c) \in D$ be the positive word and context pairs and $(w,c) \in D'$ be the negative word and context pairs. Then, using binary cross entropy loss, we learn a parameterization $\theta$ of the neural network that maximizes the function

\begin{align*}
   \mathcal{J}(\theta) &=  \sum_{(w, c) \in D} log(\sigma(v_{c}\cdot v_{w})) \\
   &+ \sum_{(w, c) \in D'} log(\sigma(-v_{c} \cdot v_{w}))
\end{align*}%
where $\sigma$ is the sigmoid function and  $v_{w}$ and $v_{c}$ are the corresponding real-valued embeddings for the target words and context words  \cite{goldberg2014word2vec}.

\begin{table*}
\centering
\begin{tabular}{ccccclllll|}
\cline{2-10}
\multicolumn{1}{l}{}             & \multicolumn{4}{|c}{McRae}                                                                     &  & \multicolumn{4}{c|}{CSLB}                                                                                           \\ \hline
\multicolumn{1}{|c|}{Model}            & \multicolumn{1}{c}{Top 1} & \multicolumn{1}{c}{Top 5} & \multicolumn{1}{c}{Top 10} & Top 20 &  & \multicolumn{1}{l}{Top 1} & \multicolumn{1}{l}{Top 5} & \multicolumn{1}{l}{Top 10} & \multicolumn{1}{l|}{Top 20} \\ \hline
\multicolumn{1}{|c|}{PLSR 50}        & 3.55                       & 14.18                      & 29.79                       & 47.52  &  & 2.90                      & 23.19                      & 44.20                       & 60.87                       \\ 
\multicolumn{1}{|c|}{PLSR 120}        & 3.55                       & 25.53                     & 43.26                       & 53.90  &  & 7.25                       & 34.78                      & 55.80                       & 71.74                       \\  \hline
\\ \hline
\multicolumn{1}{|c|}{Feature2Vec} & 4.96                      & 34.75                      & 45.39                       & 60.99  &  & 4.35                      & 36.23                      & 53.62                       & 76.09                       \\  \hline
\end{tabular}
\caption{Accuracy scores (percent) for retrieving the correct concept from amongst the top $N$ most similar nearest neighbours. For PLSR, predicted feature vectors for test concepts are compared to actual feature vectors. For Feature2Vec, predicted concept word embeddings are compared to the actual GloVe embedding vectors.}
\label{ntable}
\end{table*}

In this work, we adapt this skip-gram approach to the task of constructing semantic representations of human property norms by mapping properties into an embedding space (Figure \ref{w2v}). We achieve this by using a neural network to predict the properties from the input word, using the skip-gram architecture with negative sampling on the properties. We replace context embeddings and windowed co-occurrence counts from the conventional skip-gram architecture with property embeddings and concept-feature production frequencies. The loss function for training remains the same; however, there are two modifications to the learning process. The first is that the target embeddings for the concept words are pre-trained (i.e. the GloVe embeddings), and gradients are not applied to this embedding matrix. The layer for the property norms is randomly initialized, and gradients are applied to these vectors to learn a semantic representation for properties aligned to the pre-trained distributional semantic space for the words. Secondly, the negative samples are generated from randomly sampled properties. We downweight negative samples by multiplying their associated loss by one over the negative sampling rate, so that the system pays more attention to real cases and less to the incorrect negative examples. Due to the sparsity of word-feature production frequencies, we generate all positive instances and randomly sample negative examples after each epoch to create a new set of training samples. We name this approach \textit{Feature2Vec}\footnote{The code for the model is available at https://github.com/stevend94/Feature2Vec}. We use a learning rate of $0.001$ and Adam optimization to train for $120$ epochs. We use a negative sampling rate of $20$ for both the McRae and CLSB datasets. In contrast to the PSLR approach, \textit{Feature2Vec} learns a mapping from the feature norms onto the GloVe embedding space.

\section{Experiments}
We train with the McRae and CSLB property norms separately and report evaluations for each dataset. For the McRae dataset we use $400$ randomly selected concepts for training and the remaining $141$ for testing, and for the CSLB dataset we use $500$ randomly selected concepts for training and the remaining $138$  for testing%
\footnote{We use the Python pickle package to store the numpy state for reproducible results in our code.}.

\subsection{Predicting Feature Vectors}
We first evaluate how well the baseline PLSR model performs on the feature vector reconstruction task used by \newcite{fagarasan2015distributional}. In this evaluation, the feature vector for a test concept is predicted and we test whether the real concept vector is within the top $N$ most similar neighbours of the predicted vector. We report results over both $50$ (as in \newcite{fagarasan2015distributional}) and $120$ dimensions for a range of values of $N$ (Table \ref{ntable}). 

\subsection{Constructing Concept Representations from Feature Vectors }
For \textit{Feature2Vec}, we embed property norm features into the GloVe semantic space, giving a representation of properties in terms of GloVe dimensions. To predict a held-out concept embedding,  we build a representation of the concept word by averaging the learned feature embedding vectors for that word using the ground truth information from the property norm dataset. This gives a method to construct embeddings for new words using property knowledge and associated production frequencies (for example, for a held-out word \emph{unicorn}, its GloVe embedding vector might be predicted from all features of \emph{horse}, along with the features \emph{is-white}, \emph{has-a-horn}, and \emph{is-fantastical}). We compare these predicted embeddings to the held-out Glove embeddings (Table \ref{ntable}). However, we note that this approach is different to the PLSR models, so we do not make a direct comparison between PLSR and Feature2vec nearest neighbour results. Nevertheless, the results show that the word embeddings composed from the learned \textit{Feature2Vec} feature embeddings appear relatively frequently amongst the most similar neighbour words in the pretrained GloVe space, indicating that feature embedding composition approximates the original word embedddings reasonably well.

\begin{table}[]
\resizebox{0.48\textwidth}{!}{%
\begin{tabular}{c|ccccc|}
\cline{2-6}
                                  & \multicolumn{2}{c}{McRae}                             & \multicolumn{1}{c}{\textbf{}} & \multicolumn{2}{c|}{CSLB}                              \\ \hline
\multicolumn{1}{|c|}{Model}            & \multicolumn{1}{c}{Train} & \multicolumn{1}{c}{Test} & \multicolumn{1}{c}{\textbf{}} & \multicolumn{1}{c}{Train} & \multicolumn{1}{c|}{Test} \\ \hline
\multicolumn{1}{|c|}{PLSR 50}        & 49.52                      & 31.67                     &                                & 50.58                      & 40.25                     \\ 
\multicolumn{1}{|c|}{PLSR 120}        & 68.66                      & 32.97                     &                                & 65.42                     & 40.71                     \\ 
\multicolumn{1}{|c|}{Feature2Vec} & 90.70                      & 35.33                     &                                & 90.31                      & 44.30                     \\ \cline{1-1} \hline
\end{tabular}
}
\caption{The average percentage of features that each method can predict for a given concept vector.}
\label{scores}
\end{table}

\begin{table*}
\centering

\begin{tabular}{cc}
\hline

\textbf{Concepts}            & \textbf{Properties}            \\ 
             \hline
\multirow{3}{*}{Kingfisher} & \multirow{3}{*}{\begin{tabular}[c]{@{}c@{}}
\underline{\textbf{has\_wings}} \,
\underline{\textbf{does\_fly}} \,
\underline{\textbf{has\_a\_beak}} \,
\underline{\textbf{has\_feathers}} \,
\underline{\textbf{is\_a\_bird}} \\
\underline{\textbf{does\_eat}} \,
has\_a\_tail \,
does\_swim \,
\underline{\textbf{has\_legs}} \,
does\_lay\_eggs 
\end{tabular}} \\ & \\ &  \\ \hline
                            
\multirow{3}{*}{Avocado}    & \multirow{3}{*}{\begin{tabular}[c]{@{}c@{}}
\underline{\textbf{is\_eaten\_edible}} \,
\underline{\textbf{is\_tasty}} \,
\underline{\textbf{does\_grow}} \,
\underline{\textbf{is\_green}} \,
\underline{\textbf{is\_healthy}} \,
is\_used\_in\_cooking \\
\underline{\textbf{has\_skin\_peel}} \,
is\_red \,
is\_food \,
\underline{\textbf{is\_a\_vegetable}} \,
\end{tabular}} \\
                            &                       \\
                            &                       \\ \hline
\multirow{3}{*}{Door}                        &  \multirow{3}{*}{\begin{tabular}[c]{@{}c@{}}
made\_of\_metal  \,
has\_a\_door\_doors \, 
is\_useful  \,
has\_a\_handle\_handles  \,
made\_of\_wood  \\
made\_of\_plastic  \,
is\_heavy  \,
is\_furniture  \,
does\_contain\_hold \,
is\_found\_in\_kitchens  \, 
\end{tabular}}                   \\ 
                            &                       \\
                            &                       \\ \hline
                            
\multirow{3}{*}{Dragon}                       & \multirow{3}{*}{\begin{tabular}[c]{@{}c@{}} 
is\_big\_large  \,
is\_an\_animal  \, 
has\_a\_tail  \, 
does\_eat  \,
is\_dangerous  \\
has\_legs \, 
has\_claws  \,
is\_grey  \,
is\_small  \,
does\_fly  \,
\end{tabular}}                \\ 
                            &                       \\
                            &                       \\ \hline
\end{tabular}

\caption{Top 10 predictions from CSLB-trained Feature2Vec model. The properties in bold and underlined are those that agree with the available ground truth features from the CSLB dataset. The first two concepts are from the CSLB test set, whilst the final two words were randomly sampled from the word embedding lexicon.}
\label{featpreds}
\end{table*}

\subsection{Predicting Property Knowledge}
The evaluation task that we are most interested in is how well the models can predict feature knowledge for concepts, given the distributional semantic vectors. More specifically, for a given concept with $K$ features, we wish to take the top $K$ predicted features according to each method, and record the overlap with the true property norm listing. In this evaluation, we make direct comparisons between all three models (PLSR 50, PLSR 120, \& Feature2Vec). For the PLSR models, we predict the feature vector for a given target word using the embedding vector as input and take the top $K$ weighted features. For Feature2Vec, we rank all feature embeddings by their distance to the embedding for the target word, using cosine similarity, and take the top $K$ most similar features (Table \ref{scores}).
The results demonstrate that Feature2Vec outperforms the PLSR models on property knowledge prediction, for both training and testing datasets.

\subsection{Analysis}
Following previous work, we provide the top $10$ feature predictions for a few sample concepts, displayed in Table \ref{featpreds}. Properties underlined and in bold represent features that match the available ground truth data (i.e., the concept$\times$feature pair occurs in the norms). The first two words in Table \ref{featpreds} were sampled from the CSLB norms test set, whilst the last two words were randomly sampled from the word embedding lexicon and are not concept words appearing in the CSLB norms. We find that the predicted features that are not contained within the ground truth property set still tend to be quite reasonable, even for the two concepts not in the test dataset. As property norms do not represent an exhaustive listing of property knowledge, this is not surprising, and predicted properties not in the norms are not necessarily errors \cite{devereux2009, fagarasan2015distributional}. Moreover, the set of features used within the norms are dependent on the concepts that were presented to the human participants. It is therefore notable that the conceptual representations predicted by our model for the two out-of-norms concept words are particularly plausible, even though the attributes were never intended to conceptually represent these words. Our analysis supports the view that such supervised models could be utilised as an assistive tool for surveying much larger vocabularies of words.

\section{Conclusion}
We proposed a method for constructing distributional semantic vectors for human property norms from a pretrained vector space model of word meaning, which outperforms previous methods for predicting concept features on two property norm datasets. As discussed by \newcite{fagarasan2015distributional} and others, it is clear  that property norm datasets provide only a semi-complete picture of human conceptual knowledge, and more extensive surveys may provide additional useful property knowledge information. By predicting plausible semantic features for concepts through the leveraging of corpus-derived word embedding data, our method offers a useful tool for guiding the expensive and laborious process of collecting property norm listings. For example, existing property norm datasets can be extended through human verification of features predicted with high confidence by Feature2Vec, with these features being added to the norms and subsequently incorporated into Feature2Vec in an iterative, semi-supervised manner \cite{kelly2012}. Thus, Feature2Vec provides a useful heuristic to add interpretable feature-based information to these datasets for new words in a practical and efficient way. 

\bibliographystyle{acl_natbib_nourl}

\begin{thebibliography}{21}
\expandafter\ifx\csname natexlab\endcsname\relax\def\natexlab#1{#1}\fi

\bibitem[{Baroni and Lenci(2010)}]{baroni2010distributional}
Marco Baroni and Alessandro Lenci. 2010.
\newblock Distributional memory: A general framework for corpus-based
  semantics.
\newblock \emph{Computational Linguistics}, 36(4):673--721.

\bibitem[{Bojanowski et~al.(2017)Bojanowski, Grave, Joulin, and
  Mikolov}]{bojanowski2017enriching}
Piotr Bojanowski, Edouard Grave, Armand Joulin, and Tomas Mikolov. 2017.
\newblock Enriching word vectors with subword information.
\newblock \emph{Transactions of the Association for Computational Linguistics},
  5:135--146.

\bibitem[{Collell and Moens(2016)}]{collell2016image}
Guillem Collell and Marie-Francine Moens. 2016.
\newblock Is an image worth more than a thousand words? on the fine-grain
  semantic differences between visual and linguistic representations.
\newblock In \emph{Proceedings of COLING 2016, the 26th International
  Conference on Computational Linguistics: Technical Papers}, pages 2807--2817.

\bibitem[{Derby et~al.(2018)Derby, Miller, Murphy, and
  Devereux}]{derby2018using}
Steven Derby, Paul Miller, Brian Murphy, and Barry Devereux. 2018.
\newblock \href {https://doi.org/10.18653/v1/K18-1026} {Using sparse semantic
  embeddings learned from multimodal text and image data to model human
  conceptual knowledge}.
\newblock In \emph{Proceedings of the 22nd Conference on Computational Natural
  Language Learning}, pages 260--270, Brussels, Belgium. Association for
  Computational Linguistics.

\bibitem[{Devereux et~al.(2009)Devereux, Pilkington, Poibeau, and
  Korhonen}]{devereux2009}
Barry Devereux, Nicholas Pilkington, Thierry Poibeau, and Anna Korhonen. 2009.
\newblock \href {https://doi.org/10.1007/s11168-010-9068-8} {Towards
  unrestricted, large-scale acquisition of feature-based conceptual
  representations from corpus data}.
\newblock \emph{Research on Language and Computation}, 7(2):137--170.

\bibitem[{Devereux et~al.(2014)Devereux, Tyler, Geertzen, and
  Randall}]{devereux2014centre}
Barry~J Devereux, Lorraine~K Tyler, Jeroen Geertzen, and Billi Randall. 2014.
\newblock The {C}entre for {S}peech, {L}anguage and the {B}rain ({CSLB})
  concept property norms.
\newblock \emph{{B}ehavior {R}esearch {M}ethods}, 46(4):1119--1127.

\bibitem[{Fagarasan et~al.(2015)Fagarasan, Vecchi, and
  Clark}]{fagarasan2015distributional}
Luana Fagarasan, Eva~Maria Vecchi, and Stephen Clark. 2015.
\newblock From distributional semantics to feature norms: grounding semantic
  models in human perceptual data.
\newblock In \emph{Proceedings of the 11th International Conference on
  Computational Semantics}, pages 52--57.

\bibitem[{Goldberg and Levy(2014)}]{goldberg2014word2vec}
Yoav Goldberg and Omer Levy. 2014.
\newblock word2vec explained: deriving mikolov et al.'s negative-sampling
  word-embedding method.
\newblock \emph{arXiv preprint arXiv:1402.3722}.

\bibitem[{Kelly et~al.(2012)Kelly, Devereux, and Korhonen}]{kelly2012}
Colin Kelly, Barry Devereux, and Anna Korhonen. 2012.
\newblock \href {https://www.aclweb.org/anthology/W12-1702} {Semi-supervised
  learning for automatic conceptual property extraction}.
\newblock In \emph{Proceedings of the 3rd Workshop on Cognitive Modeling and
  Computational Linguistics ({CMCL} 2012)}, pages 11--20, Montr{\'e}al, Canada.
  Association for Computational Linguistics.

\bibitem[{Kelly et~al.(2014)Kelly, Devereux, and Korhonen}]{kelly2014}
Colin Kelly, Barry Devereux, and Anna Korhonen. 2014.
\newblock \href {https://doi.org/10.1111/cogs.12091} {Automatic extraction of
  property norm-like data from large text corpora}.
\newblock \emph{Cognitive Science}, 38(4):638--682.

\bibitem[{Levy and Goldberg(2014)}]{levy2014dependency}
Omer Levy and Yoav Goldberg. 2014.
\newblock Dependency-based word embeddings.
\newblock In \emph{Proceedings of the 52nd Annual Meeting of the Association
  for Computational Linguistics (Volume 2: Short Papers)}, volume~2, pages
  302--308.

\bibitem[{Li and Summers-Stay(2019)}]{li2019mapping}
Dandan Li and Douglas Summers-Stay. 2019.
\newblock Mapping distributional semantics to property norms with deep neural
  networks.
\newblock \emph{Big Data and Cognitive Computing}, 3(2):30.

\bibitem[{Li and Gauthier(2017)}]{lucy2017distributional}
Lucy Li and Jon Gauthier. 2017.
\newblock \href {https://doi.org/10.18653/v1/W17-2810} {Are distributional
  representations ready for the real world? {E}valuating word vectors for
  grounded perceptual meaning}.
\newblock In \emph{Proceedings of the First Workshop on Language Grounding for
  Robotics}, pages 76--85, Vancouver, Canada. Association for Computational
  Linguistics.

\bibitem[{McRae et~al.(2005)McRae, Cree, Seidenberg, and
  McNorgan}]{mcrae2005semantic}
Ken McRae, George~S Cree, Mark~S Seidenberg, and Chris McNorgan. 2005.
\newblock Semantic feature production norms for a large set of living and
  nonliving things.
\newblock \emph{Behavior {R}esearch {M}ethods}, 37(4):547--559.

\bibitem[{Mikolov et~al.(2013)Mikolov, Sutskever, Chen, Corrado, and
  Dean}]{mikolov2013distributed}
Tomas Mikolov, Ilya Sutskever, Kai Chen, Greg~S Corrado, and Jeff Dean. 2013.
\newblock Distributed representations of words and phrases and their
  compositionality.
\newblock In \emph{Advances in Neural Information Processing Systems}, pages
  3111--3119.

\bibitem[{Murphy et~al.(2012)Murphy, Talukdar, and
  Mitchell}]{murphy2012learning}
Brian Murphy, Partha Talukdar, and Tom Mitchell. 2012.
\newblock \href {https://www.aclweb.org/anthology/C12-1118} {Learning effective
  and interpretable semantic models using non-negative sparse embedding}.
\newblock In \emph{Proceedings of {COLING} 2012}, pages 1933--1950, Mumbai,
  India. The COLING 2012 Organizing Committee.

\bibitem[{Pennington et~al.(2014)Pennington, Socher, and
  Manning}]{pennington2014glove}
Jeffrey Pennington, Richard Socher, and Christopher Manning. 2014.
\newblock \href {https://doi.org/10.3115/v1/D14-1162} {{G}love: Global vectors
  for word representation}.
\newblock In \emph{Proceedings of the 2014 Conference on Empirical Methods in
  Natural Language Processing ({EMNLP})}, pages 1532--1543, Doha, Qatar.
  Association for Computational Linguistics.

\bibitem[{Rubinstein et~al.(2015)Rubinstein, Levi, Schwartz, and
  Rappoport}]{rubinstein2015well}
Dana Rubinstein, Effi Levi, Roy Schwartz, and Ari Rappoport. 2015.
\newblock How well do distributional models capture different types of semantic
  knowledge?
\newblock In \emph{Proceedings of the 53rd Annual Meeting of the Association
  for Computational Linguistics and the 7th International Joint Conference on
  Natural Language Processing (Volume 2: Short Papers)}, volume~2, pages
  726--730.

\bibitem[{{\c{S}enel} et~al.(2018){\c{S}enel}, {Utlu}, {Y\"ucesoy}, {Ko\c{c}},
  and {\c{C}ukur}}]{senel2018}
L.~K. {\c{S}enel}, I.~{Utlu}, V.~{Y\"ucesoy}, A.~{Ko\c{c}}, and T.~{\c{C}ukur}.
  2018.
\newblock \href {https://doi.org/10.1109/TASLP.2018.2837384} {Semantic
  structure and interpretability of word embeddings}.
\newblock \emph{IEEE/ACM Transactions on Audio, Speech, and Language
  Processing}, 26(10):1769--1779.

\bibitem[{Silberer(2017)}]{silberer2017grounding}
Carina Silberer. 2017.
\newblock Grounding the meaning of words with visual attributes.
\newblock In \emph{Visual {A}ttributes}, pages 331--362. Springer.

\bibitem[{Sommerauer and Fokkens(2018)}]{sommerauer2018firearms}
Pia Sommerauer and Antske Fokkens. 2018.
\newblock \href {https://doi.org/10.18653/v1/W18-5430} {Firearms and tigers are
  dangerous, kitchen knives and zebras are not: Testing whether word embeddings
  can tell}.
\newblock In \emph{Proceedings of the 2018 {EMNLP} Workshop {B}lackbox{NLP}:
  Analyzing and Interpreting Neural Networks for {NLP}}, pages 276--286,
  Brussels, Belgium. Association for Computational Linguistics.

\end{thebibliography}

\end{document}